\documentclass[runningheads]{llncs}
\usepackage[T1]{fontenc}
\usepackage{fmtcount}
\usepackage[colorlinks,linkcolor=blue,urlcolor=blue]{hyperref}
\usepackage{graphicx,verbatim}
\usepackage{amssymb}
\usepackage{amsmath}
\usepackage{multirow}
\usepackage{tabularx}
\usepackage{gensymb}
\usepackage{url}
\usepackage[table,xcdraw]{xcolor}
\usepackage{xcolor}
\usepackage{graphicx} 
\usepackage{booktabs}
\usepackage{mathrsfs}
\usepackage{pifont}
\usepackage[misc]{ifsym}

\begin{document}
\title{Uncertainty-aware Diffusion and Reinforcement Learning for Joint Plane Localization and Anomaly Diagnosis in 3D Ultrasound}
\titlerunning{Joint Plane Localization and Anomaly Diagnosis in 3D Ultrasound}

\author{
Yuhao Huang\inst{1} \and 
Yueyue Xu\inst{2} \and 
Haoran Dou\inst{3} \and 
Jiaxiao Deng\inst{1} \and 
Xin Yang\inst{1} \and 
\\ Hongyu Zheng\inst{2} \and 
Dong Ni\inst{1,4,5,6}\textsuperscript{(\Letter)} 
} 
\authorrunning{Y. Huang et al.}

\institute{
Medical Ultrasound Image Computing (MUSIC) Lab, School of Biomedical Engineering, Medical School, Shenzhen University, Shenzhen, China\\
\email{nidong@szu.edu.cn} \\ \and
The People’s Hospital of Guangxi Zhuang Autonomous Region, Nanning, China\\ \and
Centre for CIMIM, Manchester University, Manchester, UK\\ \and
School of Artificial Intelligence, Shenzhen University, Shenzhen, China\\ \and
National Engineering Laboratory for Big Data System Computing Technology, Shenzhen University, Shenzhen, China\\ \and
School of Biomedical Engineering and Informatics, Nanjing Medical University, Nanjing, China}

\maketitle             

\begin{abstract}
Congenital uterine anomalies (CUAs) can lead to infertility, miscarriage, preterm birth, and an increased risk of pregnancy complications. Compared to traditional 2D ultrasound (US), 3D US can reconstruct the coronal plane, providing a clear visualization of the uterine morphology for assessing CUAs accurately. In this paper, we propose an intelligent system for simultaneous automated plane localization and CUA diagnosis. Our highlights are: 1) we develop a denoising diffusion model with local (plane) and global (volume/text) guidance, using an adaptive weighting strategy to optimize attention allocation to different conditions; 2) we introduce a reinforcement learning-based framework with unsupervised rewards to extract the key slice summary from redundant sequences, fully integrating information across multiple planes to reduce learning difficulty; 3) we provide text-driven uncertainty modeling for coarse prediction, and leverage it to adjust the classification probability for overall performance improvement. Extensive experiments on a large 3D uterine US dataset show the efficacy of our method, in terms of plane localization and CUA diagnosis. Code is available at \href{https://github.com/yuhoo0302/CUA-US}{GitHub}.

\end{abstract}
%
%
%
\section{Introduction}
\label{sec:intro}
Congenital uterine anomalies (CUAs) are one of the leading causes of infertility, abnormal fetal positioning, and preterm birth~\cite{chan2011prevalence}.
2D ultrasound (US) is commonly used for initial anomaly screening, but its limited spatial information may lead to misdiagnosis.
In comparison, 3D US provides a more detailed visualization of uterine morphology and the structural spatial relationship~\cite{kougioumtsidou2019three}.
Moreover, it can reconstruct the coronal plane, which is infeasible for 2D US in practice, to show clear uterine fundus and cavity features for improving diagnostic accuracy.
However, manually localizing the standard plane and performing efficient diagnosis is challenging and biased due to the vast search space, orientation variability, anatomical complexity and internal invisibility of 3D US.
Thus, developing an automatic approach to address both tasks simultaneously is essential to alleviate the burden on sonographers and reduce operator dependency.

\textbf{Plane localization in 3D medical images.}
Compared with the one-step method that directly transforms the volume data into plane parameters, recent iterative methods have shown better localization performance~\cite{li2018standard}.
Several reinforcement learning (RL) based methods~\cite{alansary2018automatic,yang2021agent,yang2021searching,zou2022agent,huang2023localizing}, equipped with different strategies, including registration, early-stop termination, neural architecture search, and tangent-based formulation, have achieved satisfactory localization accuracy in 3D US.
Most recently, Dou et al.~\cite{dou2025standard} proposed a denoising diffusion model with spherical tangent to drive a continuous and robust transformation.
They also explored the localization inconsistency score to perform threshold-based binary classification (normal/abnormal) and obtain 90.67\% accuracy.

\textbf{Deep learning-based 3D medical classification.} 
Previous studies have adopted different techniques, including multi-task learning~\cite{yang20243d}, multi-plane attention~\cite{jang2022m3t}, and long-range modeling~\cite{gong2025nnmamba}, for 3D classification.
Yang et al.~\cite{yang2023medmnist} introduced the MedMNIST v2 dataset and benchmarked different deep models.
Currently, although there are limited studies directly targeting 3D US, various US video classification approaches~\cite{sun2023boosting,wang2022key} achieved good performance on different tasks and have the potential to handle 3D tasks.
However, most existing methods relied only on global volumetric knowledge or local information from slices/planes.
Moreover, they only output the final classification categories without any intermediate results, limiting their interpretability and clinical availability.
Thus, they may not suit our CUA diagnosis task.

In this work, we propose a joint framework based on uncertainty-aware diffusion and RL to automatically localize the coronal plane and diagnose CUAs in 3D US. 
We believe this is the first work to integrate these two essential tasks in one unified framework, matching the clinical workflow in uterine examinations.
Our contribution is three-fold.
First, we adopt adaptive multi-scale conditions to guide the denoising process in the diffusion model for accurate plane detection.
Second, we leverage the RL agent with unsupervised rewards to provide slice summary, balancing learning difficulty and information preservation.
Third, we utilize the uncertainty score powered by text conditions to refine the original classification probability and further improve performance.
Validated on the large uterine US dataset demonstrates that our method outperforms strong competitors in both localization and diagnosis tasks, showing satisfactory performance.

\section{Methodology}
\label{sec:Method}
\begin{figure}[!t]
\centering
\includegraphics[width=1.0\textwidth]{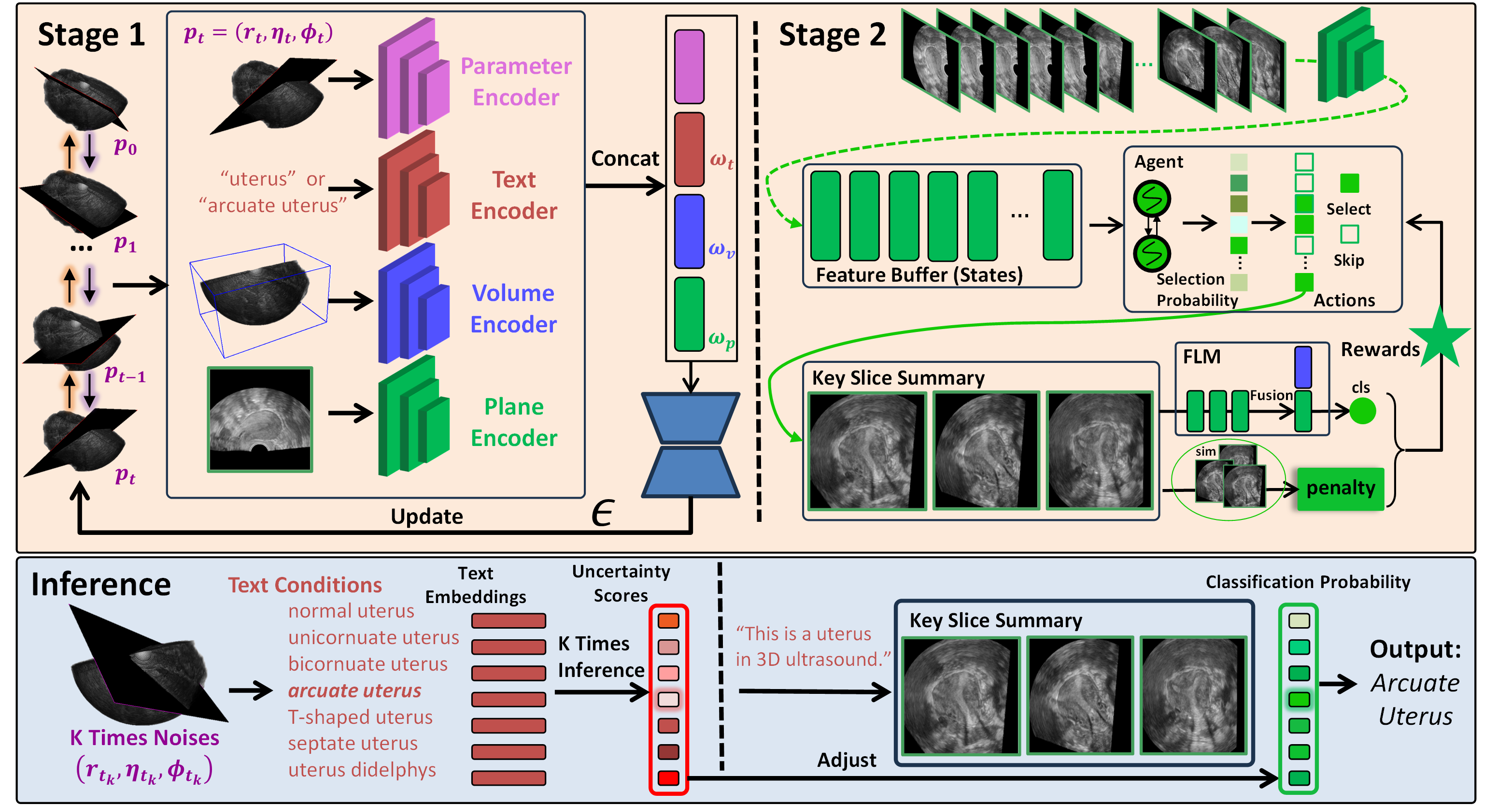}
\caption{Overview of our proposed framework.} 
\label{fig:framework}
\end{figure}

Fig.~\ref{fig:framework} shows the schematic view of our proposed method.
In stage 1, we first add \textit{Gaussian} noise to the target plane parameter $p_0$, and adaptively perform noisy plane refinement conditioned on multi-scale guidance (i.e., plane, volume and text).
In stage 2, we leverage the RL strategy to select the key slices from the above iterative process to build the slice summary and boost model learning.
During inference, we obtain the text-driven uncertainty scores to adjust the original classification probabilities, finally improving the overall performance.

\subsection{Adaptive Conditional Diffusion Model for Plane Localization}
To achieve accurate and efficient plane localization in 3D US, inspired by~\cite{dou2025standard}, we formulate this task as a conditional generative problem using the diffusion model~\cite{song2020denoising}.
The plane function is represented using tangent-point in spherical coordinates~\cite{dou2025standard}, i.e., $p=(r,\eta,\theta)$, to alleviate the angular sensitivity.
Give a SP parameter $p_0$, during the forward diffusion process with \textit{t} steps, $p_t$ can be defined as $\sqrt{\alpha_t} p_{0} + \sqrt{1 - \alpha_t} \epsilon$, where $\alpha_t=\prod_{s=1}^t (1 - \beta_s)$ with $\beta_s$ is the noise schedule hyperparameter.
Then, the reverse denoising process can be written as:
\begin{equation}
p_{t-1} = \sqrt{\alpha_{t-1}} \left(\frac{p_t - \sqrt{1-\alpha_t} \epsilon_\theta(p_t, t, c)}{\sqrt{\alpha_t}}\right) + \sqrt{1-\alpha_{t-1}} \epsilon_\theta(p_t, t, c),
\end{equation}
where $c$ is the condition set including volume and plane ($c_v$, $c_p$) features by convolution encoders, and text embeddings $c_t$ from BiomedCLIP~\cite{zhang2023biomedclip}.
Texts aim to equip the localization model with semantic information for following classification~\cite{clark2024text}.
Specifically, during training, we randomly provide text prompts like `This is a uterus in 3D ultrasound' or `This is a $\mathcal{U}_{gt}$ uterus in 3D ultrasound'.
$\mathcal{U}_{gt}$ represents the ground truth (GT) of CUA category.
We also introduce the adaptive parameters to control the weights of different conditions.
The features from different encoders and the timestep embeddings were first concatenated and mapped to the weights ($\omega_{v},\omega_{p},\omega_{t}$) using linear projections with $Sigmoid$.
Then, the training loss can be defined as:
$\mathcal{L} = \mathbb{E}_{x_0, c, t, \epsilon \sim \mathcal{N}(0, 1)} \left[ \| \epsilon - \epsilon_\theta(x_t, t, c(\omega) \|^2 \right]$.

During inference, we provide the category-free text `This is a uterus ultrasound image'.
Then, \textit{K=8} randomly noise plane parameters $p_k$ are generated as initial inputs.
After the denoising process, \textit{K} predicted parameters are obtained, and the final predicted plane can be determined by the averaged parameter.

\subsection{RL with Unsupervised Reward for Key Slice Summary}
The denoising process will output a series of 2D planes $s_1,s_2,...s_T$ with features $f_1,f_2,...f_T$. 
These bring vital local information to perform classification.
However, not all slices can contribute positively, since some may contain diagnosis-irrelevant features.
Moreover, using only the final slice may seriously lose vital local information embedded in the denoising process.
Hence, an algorithm that can select key slices should be designed to discard unimportant features, preserve slices of fruitful anatomical knowledge, and improve classification accuracy.

Inspired by~\cite{huang2022extracting}, we introduce an RL-based solution to extract representative slices and remove redundant information.
Specifically, we input $T$ features from the buffer into the designed RL agent and obtain the plane-wise selection probability $prob_t$.
The agent includes a Bi-LSTM and a fully connected layer with \textit{sigmoid}.
Then, the action $a_t\in \{0,1\}$ can be sampled by $Bernoulli(prob_t)$, where 1 indicates that the $t^{th}$ plane should be selected.
A key slice set $S$ can be constructed by selecting the final prediction and planes with $a_t=1$.
For evaluating the quality of $\mathcal{S}$, a reward function is required to balance the redundancy and retention of anatomical information while controlling the set size ($|\mathcal{S}|$):
\begin{equation}
R = \underbrace{\frac{1}{|\mathcal{S}|(|\mathcal{S}|-1)} \sum_{i \neq j} sim(f_i, f_j)}_{R_{sim}} + \underbrace{\sum_{c=1}^{C} y_c log(\hat{y}_c)}_{R_{cls}} - \underbrace{\alpha \cdot max(0, e^{\gamma(|\mathcal{S}|-\mathcal{S}_{max})})}_{R_{penalty}},
\end{equation}
where $R_{sim}$ calculates the cosine similarities ($sim$) between features to constrain redundancy, $R_{cls}$ uses the classification loss to feedback anatomical retention ($C$ and $y_c$ represents the category and the corresponding probabilities).
Specifically, by taking the maximum value at the time dimension, the features of key slices of size ($\mathcal{S},\mathcal{D}$) will be fused to $\hat{f}_{\mathcal{S}}=(1,\mathcal{D})$, which then be combined with the volume features and inputted into a linear layer with \textit{Softmax} for classification output (named fusion linear module, FLM).
$R_{penalty}$ penalizes the situations of storing over $\mathcal{S}_{max}=5$ slices.
Since no key slice labels are available, the reward functions are conducted in an unsupervised manner.
Our goal is to learn a policy ($\pi_\zeta$) that maximizes the expected reward $J(\zeta)$.
We follow REINFORCE algorithm~\cite{williams1992simple} and approximate the gradient by running $N$=100 episodes and averaging the results:
\begin{equation}
\nabla_{\zeta} J(\zeta) \approx \frac{1}{N} \sum_{i=1}^{N} \sum_{t=1}^{T} (R_n-b) \nabla_{\zeta} \log \pi_{\zeta}(a_t | f_t(s_t)),
\end{equation}
where $R_n$ is the reward at $n$ episode and $b$ is the moving average of previous rewards to accelerate convergence.
Then, with the L2 regularization, we update $\zeta$ by $\zeta_{i+1}=\zeta_i-\rho\nabla_{\zeta}(\eta\sum_{i,j}\zeta_{i,j}^{2}-J(\zeta))$.
Moreover, we optimize the FLM by cross-entropy loss.
To ease learning, the training of the agent and FLM alternates.

\subsection{Uncertainty-aware Strategy for Classification Adjustment}
The text-to-image diffusion model has proven its zero-shot classification ability in medical tasks~\cite{favero2025conditional}.
Specifically, classification is performed by comparing reconstruction errors across images generated for each possible condition (e.g., sick/healthy).
Here, we assume that giving the correct text condition will bring less uncertainty.
During testing, we generate texts ($c_{t_i}$) `This is a $\mathcal{U}$ in 3D ultrasound', for all uterine types ($\mathcal{U}$).
The uncertainty-aware score for $c_{t_i}$ is:
\begin{equation}
S_{unc}(c_{t_i}) = \sum_{i=1}^{K} \left\| p_k(c_{t_i}) - \frac{1}{K} \sum_{k=1}^{K} p_k(c_{t_i}) \right\|^2,
\end{equation}
where $p_k(c_{t_i})$ indicates the predicted plane parameter under $c_{t_i}$.
Then, the coarse classification result represents the category corresponding to the smallest uncertainty value, i.e., $\arg\min_{c}S_{unc}(c)$.
We also design a simple yet effective way to connect the uncertainty score and original classification probability ($p_o$).
Specifically, we first perform normalization on the uncertainty score, and obtain the uncertainty probability ($p_u$), then the adjusted probability ($p_a$) is written as:
\begin{equation}
p_{\text{a}}(y_i) = \frac{p_{o}(y_i) \cdot \frac{1}{p_{u}(y_i)}}{\sum_{j=1}^{N} (p_{o}(y_j) \cdot \frac{1}{p_u}(y_j)}.
\end{equation}

\section{Experimental Results}
\label{sec:Experimental Results}
\textbf{Materials and Implementation Details.}
Our private large 3D uterus dataset in pelvic US contains 677 volumes, annotated with coronal standard planes, and uterine categories: normal (N, 345), unicornuate (U, 52), bicornuate (B, 5), arcuate (A, 145), T-shaped (T, 15), septate uterus (S, 96), and uterus didelphys (D, 19), by experienced sonographers under strict quality control using the Pair annotation software package~\cite{liang2022sketch}.
The average volume size is 367$\times$189$\times$334, with spacing equal to 0.4$^{3}$ \textit{$mm^{3}$}.
We randomly split the volumes into 406, 68, and 203 for training, validation, and testing, respectively.

\begin{table}[!t]
  \centering 
  \caption{Comparison on plane localization task. The best results are shown in bold.} 
    \begin{tabular}{ccccccc}
    \toprule
          & ITN~\cite{li2018standard}   & RL$_{WSDT}$~\cite{yang2021agent} & RL$_{NAS}$~\cite{yang2021agent} & RL$_{FT}$~\cite{zou2022agent} & DIFF$_{MSG}$~\cite{dou2025standard} & Ours \\
    \midrule
    Ang   & 30.36±18.45 & 17.71±15.69 & 13.22±12.76 & 12.65±11.32 & 8.84±6.77 & \textbf{7.12±5.16} \\
    Dis   & 1.24±1.99 & 1.55±1.44 & 1.20±1.09 & 1.26±1.10 & 1.01±0.84 & \textbf{0.86±0.78} \\
    SSIM  & 0.57±0.16 & 0.63±0.14 & 0.71±0.11 & 0.72±0.09 & 0.74±0.13 & \textbf{0.76±0.12} \\
    NCC   & 0.61±0.14 & 0.69±0.13 & 0.76±0.13 & 0.75±0.13 & 0.80±0.19 & \textbf{0.82±0.16} \\
    \bottomrule
    \end{tabular}%
  \label{tab:plane_loc}%
\end{table}%

\begin{table}[!t]
  \centering
  \caption{Comparison on CUA classification task. The best results are shown in bold.}
  \setlength{\tabcolsep}{1.9mm}
  \begin{tabular}{ccccccccccc}
    \toprule
          & \multicolumn{5}{c}{Method}            & Acc   & Pre   & Rec   & F1    & AUC \\
    \midrule
    \multirow{3}[2]{*}{2D/2.5D} & \multicolumn{5}{c}{ResNet18~\cite{he2016deep}}          & 51.23  & 32.68  & 48.52  & 34.40  & 0.8753\\
          & \multicolumn{5}{c}{ResNet18-2.5D~\cite{he2016deep}}     & 70.44  & 45.75  & 40.10  & 35.73  & 0.9350  \\
          & \multicolumn{5}{c}{ResNet18-LSTM~\cite{he2016deep}}     & 75.86  & 63.84  & 39.98  & 39.50  & 0.9628  \\
    \midrule
    \multirow{3}[2]{*}{Video} & \multicolumn{5}{c}{I3D~\cite{carreira2017quo}}               & 70.44  & 74.52  & 59.77  & 61.54  & 0.9600 \\
          & \multicolumn{5}{c}{VST~\cite{liu2022videost}}               & 88.18  & 63.19  & 65.60  & 63.20  & 0.9912  \\
          & \multicolumn{5}{c}{UniFormerV2~\cite{li2022uniformerv2}}       & 89.16  & 77.02  & 63.65  & 64.11  & 0.9929  \\
    \midrule
    \multirow{7}[2]{*}{3D} & \multicolumn{5}{c}{ResNet18~\cite{he2016deep}}          & 74.38  & 73.62  & 58.51  & 59.43  & 0.9715 \\
          & \multicolumn{5}{c}{ResNet50~\cite{he2016deep}}          & 76.35  & 71.11  & 60.79  & 55.92  & 0.9766  \\
          & \multicolumn{5}{c}{MedicalNet18~\cite{chen2019med3d}}      & 84.24  & 73.88  & 73.54  & 70.20  & 0.9800 \\
          & \multicolumn{5}{c}{MedicalNet50~\cite{chen2019med3d}}      & 85.22  & 96.54  & 65.20  & 67.81  & 0.9806   \\
          & \multicolumn{5}{c}{DenseNet121~\cite{huang2017densely}}       & 74.88  & 89.93  & 81.26  & 78.83  & 0.9768  \\
          & \multicolumn{5}{c}{ViT~\cite{dosovitskiy2020image}}               & 76.35  & 74.84  & 74.16  & 71.34  & 0.9732 \\
          & \multicolumn{5}{c}{nnMamba~\cite{gong2025nnmamba}}           & 86.70  & 87.95  & 89.51  & 83.33  & 0.9903  \\
          \midrule
              & AP     & PP     & SS     & GF     & UA     & Acc   & Pre   & Rec   & F1    & AUC \\
    \midrule
    \multirow{6}[2]{*}{Ours} & \ding{51} & \ding{55} & \ding{55} & \ding{55} & \ding{55} & 91.13  & 97.86  & 88.42  & 91.96  & 0.9944  \\
          & \ding{51} & \ding{55} & \ding{55} & \ding{51} & \ding{55} & 92.12  & 72.78  & 74.86  & 71.46  & 0.9971  \\
          & \ding{55} & \ding{51} & \ding{55} & \ding{51} & \ding{55} & 87.19  & 71.17  & 74.41  & 68.42  & 0.9867 \\
          & \ding{55} & \ding{55} & \ding{51} & \ding{51} & \ding{55} & 92.61  & 98.17  & 91.11  & 93.91  & 0.9968  \\
          & \ding{55} & \ding{55} & \ding{55} & \ding{55} & \ding{51} & 92.12  & 97.60  & 87.26  & 91.28  & 0.9961  \\
          & \ding{55} & \ding{55} & \ding{51} & \ding{51} & \ding{51} & \textbf{94.58}  & \textbf{98.34}  & \textbf{98.76}  & \textbf{95.21}  & \textbf{0.9989}  \\
    \bottomrule
    \end{tabular}%
  \label{tab:cls1}%
\end{table}%

In this study, we implemented our method in \textit{PyTorch}, using one NVIDIA 4090 GPU with 24GB memory.
In stage 1, the denoising diffusion model with cosine noise scheduler was developed based on \textit{diffusers} package by \textit{HuggingFace}.
The time steps are set to 1000 for training, and 100 for validation and testing.
The training includes 200\textit{K} interactions with batch size=8 and learning rate ($lr$)=5e-4.
The architectures of parameter/plane/volume encoders follow the design in~\cite{dou2025standard}, here, we used different multi-layer perceptions (MLPs) to map them to $\mathbb{R}^{32}$.
The text encoder followed BiomedCLIP~\cite{zhang2023biomedclip}, obtaining $\mathbb{R}^{32}$ features via MLP.
Through concatenation, they were input into the UNet-based denoising network (with the positional encoding of timestep added to all blocks).
In stage 2, we trained the RL agent and FLM with lr=5e-4 in 100\textit{K} interactions.
Alternate training was conducted every 1\textit{K} iterations.
Both stages used AdamW optimizers and conducted validation every 2\textit{K} iterations, and models with the best validation performance were saved for final evaluation.
All slices were resized to 224$^{2}$, and volume size standardization was not required since we processed one volume per iteration.
They were then scaled to 0–1 by dividing 255.

\textbf{Quantitative and Qualitative Analysis.}
For plane localization, we analyzed the performance in terms of spatial differences (\textit{Ang}, \textit{Dis}) and content similarities (\textit{SSIM}, \textit{NCC}).
\textit{Ang} (\degree) calculates the angles between normal vectors of two planes, and \textit{Dis} ($mm$) denotes the difference between their Euclidean distances towards the volume origin.
Details for \textit{SSIM} and \textit{NCC} refer to~\cite{dou2025standard}.
For classification, Accuracy (Acc, \%), Precision (Pre, \%), Recall (Rec, \%), F1-score (F1, \%), and AUC were included for comprehensive analysis.
Besides, we used the paired t-test to assess statistical significance for plane localization metrics, and the Chi-square test for overall classification performance.

\begin{table}[!t]
  \centering
  \setlength{\tabcolsep}{1mm}
  \caption{Different localization methods with ($w$) and without ($w/o$) slice summary (SS) for classification. The improved metrics are highlighted in bold.}
    \begin{tabular}{lcccccccccc}
    \toprule
    \multirow{2}[4]{*}{} & \multicolumn{2}{c}{RL$_{WSDT}$} & \multicolumn{2}{c}{RL$_{NAS}$} & \multicolumn{2}{c}{RL$_{TF}$} & \multicolumn{2}{c}{DIFF$_{MSG}$} & \multicolumn{2}{c}{Ours} \\
\cmidrule{2-11}          & \textit{w/o} SS & $w$ SS & \textit{w/o} SS & $w$ SS & \textit{w/o} SS & $w$ SS & \textit{w/o} SS & $w$ SS & \textit{w/o} SS & $w$ SS \\
    \midrule
    Acc   & 79.81  & \textbf{81.28}  & 83.74  & \textbf{85.22}  & 86.70  & \textbf{87.68}  & 89.66  & \textbf{90.64}  & 91.13  & \textbf{93.10}  \\
    F1    & 64.74  & \textbf{69.05}  & 70.85  & \textbf{73.69}  & 74.76  & \textbf{78.71}  & 78.46  & \textbf{83.35}  & 91.96  & \textbf{93.38}  \\
    \bottomrule
    \end{tabular}%
  \label{tab:cls2}%
\end{table}%

\begin{figure}[!t]
\centering
\includegraphics[width=0.95\textwidth]{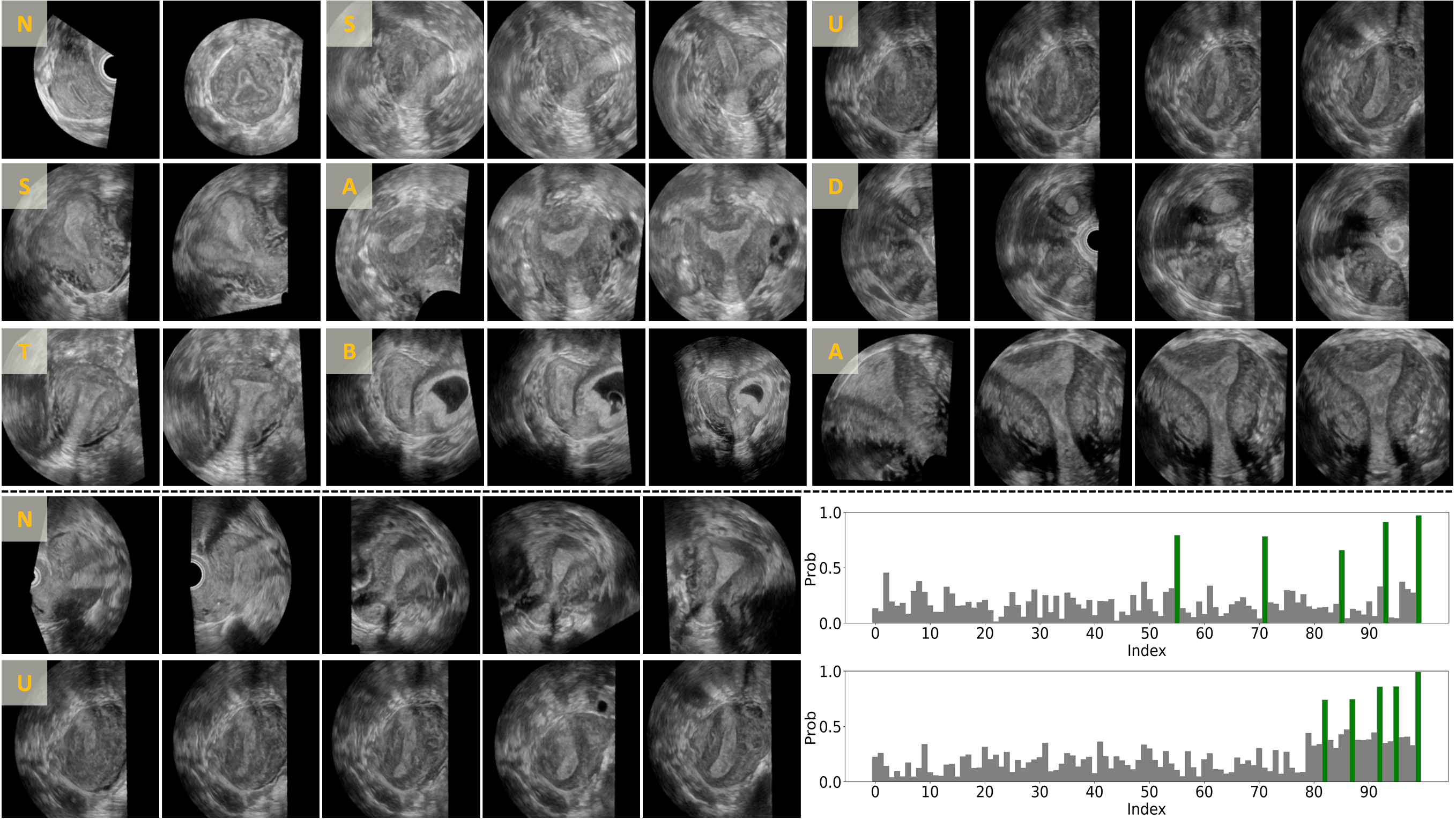}
\caption{Typical summary examples with uterine types in the top-left corners. 
Below are two summaries with index selection probabilities: green (select) and gray (skip).} 
\label{fig:result}
\end{figure}

\begin{figure}[!t]
\centering
\includegraphics[width=1.0\textwidth]{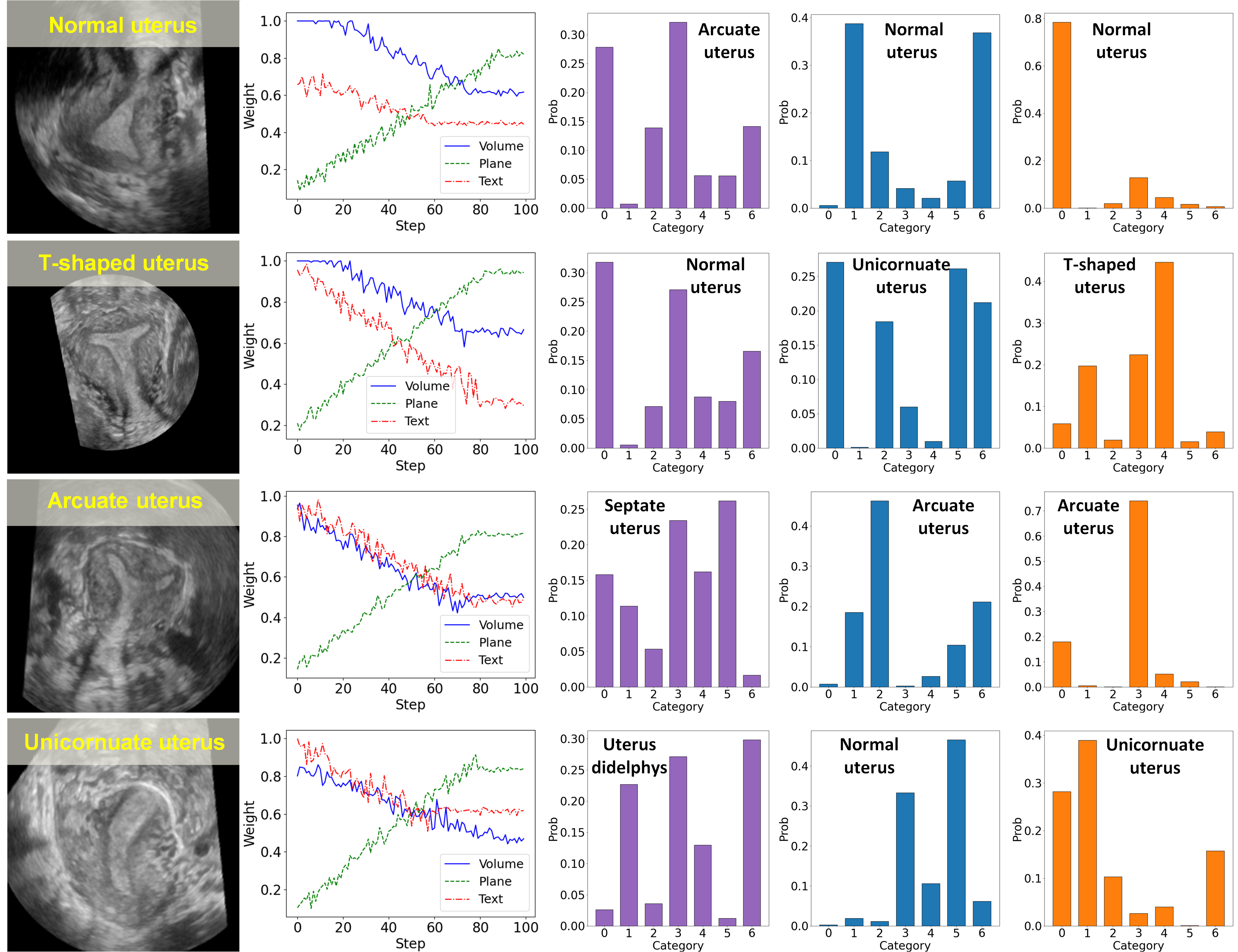}
\caption{Visualization results. 
c1: predicted planes with GTs (yellow).
c2: denoising curves of condition weights at different time steps.
c3-5: original, uncertainty and adjusted probabilities with predicted classification results (c: column).} 
\label{fig:result1}
\end{figure}

Table~\ref{tab:plane_loc} compares our method with five iterative approaches, i.e., one traditional CNN~\cite{li2018standard}, three RL methods~\cite{yang2021agent,yang2021searching,zou2022agent}, and one diffusion-based solution~\cite{dou2025standard}.
Results show that our proposed method significantly outperforms all the competitors ($p<0.05$) in terms of spatial and content similarities.
In Table~\ref{tab:cls1}, we evaluated various typical and advanced methods and their variants, containing three 2D/2.5D traditional networks~\cite{he2016deep}, three video-based approaches~\cite{carreira2017quo,liu2022videost,li2022uniformerv2}, and seven 3D models~\cite{he2016deep,chen2019med3d,huang2017densely,dosovitskiy2020image,gong2025nnmamba}.
Results show that they all obtain \textit{Acc} and \textit{F1} less than 90\%.
Our proposed method shows significant improvement over them in all evaluation metrics (last row, $p<0.05$), validating its effectiveness.
Table~\ref{tab:cls1} also provides the ablation studies to test the contribution of each proposed strategy.
$AP$, $PP$, and $SS$ represent using all planes, the final predicted plane, and slice summary as classification inputs, respectively.
$GF$ means the global volumetric features, and $UA$ denotes the uncertainty-aware strategy.
We observe that even with $GF$, using only one plane degrades the performance compared to using all planes. 
Besides, $SS$ can remove abundant information, extract key features, and improve performance (F1: 22.45\%$\uparrow$).
Notably, with $UA$ only, our model achieves satisfactory performance with Acc, F1 and Pre$>$90\%.
Adding $SS$ and $GF$ can further enhance the performance on all evaluation metrics.
We further prove that $SS$ is general to different plane localization methods, with significantly improved performance (Table~\ref{tab:cls2}, all p-values$<$0.05).

We also visualize the key slice summaries in Fig.~\ref{fig:result}.
It shows that the selected slices are representative and diverse, capturing the vital anatomical information of the corresponding CUA types.
Besides, the denoising curves in Fig.~\ref{fig:result1} prove that global conditions (volume/text) play a vital role in the early denoising stages, driving coarse localization, while local plane conditions have minimal impact initially. 
As denoising progresses, the model gradually shifts focus to local information to achieve fine plane localization.
Bar plots in the last three columns (c) show examples where raw or uncertainty-based predictions were incorrect (c3-4), and how uncertainty adjustment corrects the results (c5).

\section{Conclusion}
\label{sec:conclusion}
In this paper, we propose a novel joint framework for automated plane localization and CUA diagnosis in 3D US.
We first introduce the adaptive denoising strategy to weight conditions and improve localization.
Then, we propose RL with unsupervised rewards for key slice summary construction, thereby enhancing model learning.
Last, we leverage the text condition-driven uncertainty-aware score to adjust the classification probability and boost the overall performance.
In the future, we will extend the method to more modalities and tasks.

\begin{credits}
\subsubsection{\ackname}
This work was supported by the grant from National Natural Science Foundation of China (12326619, 62171290, 82201851); Science and Technology Planning Project of Guangdong Province (2023A0505020002); Frontier Technology Development Program of Jiangsu Province (BF2024078); Guangxi Province Science Program (2024AB17023), Key Research and Development Program (AB23026042) and Natural Science Foundation (2025GXNSFAA069471).

\subsubsection{\discintname}
The authors have no competing interests to declare that are relevant to the content of this article.
\end{credits}

\bibliographystyle{splncs04}
\bibliography{Paper-5180}

\end{document}